\documentclass[final,5p,times,twocolumn]{elsarticle}

\usepackage{amsmath,amssymb,amsfonts}
\usepackage{algorithmic}
\usepackage{graphicx}
\usepackage{textcomp}
\usepackage{orcidlink}

\usepackage{enumitem} 
\usepackage{booktabs}
\usepackage{url}
\usepackage{hyperref}

\usepackage{calrsfs}
\DeclareMathAlphabet{\pazocal}{OMS}{zplm}{m}{n}

\usepackage{lipsum}

\journal{Computer Methods and Programs in Biomedicine}

\begin{document}

\begin{frontmatter}



\title{Multi-Rater Calibrated Segmentation Models}

\author[sycai,upf]{Meritxell Riera-Marín} 
\author[sycai]{Javier García López}
\author[sycai]{Júlia Rodríguez-Comas}
\author[upf,icrea]{Miguel A. González Ballester}
\author[tecnalia]{Adrian Galdran}

\affiliation[sycai]{organization={Sycai Technologies SL, Scientific and Technical Department},
             city={Barcelona},
             country={Spain}}

\affiliation[upf]{organization={BCN Medtech, Universitat Pompeu Fabra},
             city={Barcelona},
             country={Spain}}

\affiliation[icrea]{organization={Institució Catalana de Recerca i Estudis Avançats (ICREA)},
             city={Barcelona},
             country={Spain}}
             
\affiliation[tecnalia]{organization={TECNALIA, Basque Research and Technology Alliance (BRTA)},
             city={Bizcaia},
             country={Spain}}

\begin{abstract}
\textbf{Objective:} Accurate probability estimates are essential for the safe deployment of medical image segmentation models in clinical decision-making. However, modern deep segmentation networks are often poorly calibrated, a problem exacerbated when multiple expert annotations exhibit substantial disagreement. While inter-rater variability is typically treated as noise, it provides valuable information about intrinsic annotation ambiguity that must be reflected in model confidence. \\
\textbf{Methods:} We improve the probabilistic calibration of medical image segmentation models by reformulating multi-rater supervision as an ordinal learning problem. Voxel-wise annotator agreement is treated as an ordered target, linking predictive confidence to the empirical variability in training data. This formulation allows the use of ordinal-aware scoring rules, such as the Ranked Probability Score ordinal loss, combined with a standard binary objective to preserve discriminative performance. \\
\textbf{Results:} We evaluated the proposed approach across four public segmentation benchmarks spanning ophthalmology, histopathology, and thoracic imaging. Calibration was assessed using a multi-rater extension of expected calibration error. Results consistently show that ordinal-aware training yields substantially improved calibration with respect to inter-rater agreement without degrading segmentation accuracy. \\
\textbf{Conclusions:} Treating multi-rater annotations as ordered information provides a principled and architecture-agnostic route to more reliable probabilistic segmentation models
\end{abstract}




\end{frontmatter}



\begin{figure*}[!t]
\centering
\includegraphics[width=0.9\textwidth]{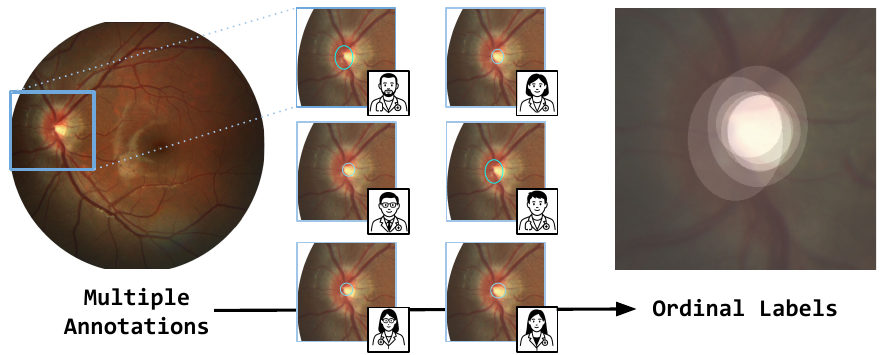} 
\caption{
Overview of the proposed Ordinal Calibration framework. 
The pipeline transforms $K$ independent binary annotations into a single ordinal ground-truth map representing the rater consensus level $L \in \{0, \dots, K\}$. The segmentation network maps input pixels to a categorical probability distribution over $K+1$ ordinal levels, supervised by a hybrid RPS-BCE loss. This formulation assigns graded penalties proportional to the ordinal distance from the empirical rater consensus, effectively regularizing model confidence in ambiguous regions. By treating rater disagreement as a structured hierarchy rather than independent noise, the framework achieves superior calibration without compromising discriminative performance.
}\label{key_figure}
\end{figure*}

\section{Introduction}\label{sec_introduction}

Precise image segmentation is a central topic in current medical image analysis, as it is crucial for early diagnosis, personalized treatment planning, and outcome prediction. 
Machine learning and artificial intelligence (AI) have significantly advanced the analysis of complex imaging data, but their trustworthiness remains an essential prerequisite for clinical adoption. In this context, model calibration has emerged as a key property of reliable predictive systems \cite{mehrtash_confidence_2020}. 
Calibration refers to the ability of a model to formulate probabilistic predictions aligned with its own accuracy, ensuring that lower predictive confidence is truly related to less likelihood of being correct. 
Well-calibrated models not only provide accurate predictions but also express appropriate confidence in their outputs. 
This is especially critical in medical decision-making, where both overconfidence and underconfidence can lead to misdiagnosis or suboptimal treatment choices, with serious clinical consequences. 
For instance, overconfidence in an incorrect segmentation of a tumor margin can directly compromise radiotherapy planning \cite{sherer_metrics_2021}, while underconfidence may cause clinicians to disregard an accurate prediction.
In practice, many diagnostic and treatment decisions rely on a model’s probabilistic estimates, meaning that poor calibration can lead either to unwarranted trust in incorrect predictions or to the underutilization of otherwise accurate models \cite{chua_tackling_2023}.

Deep neural networks inherently tend to produce overconfident probability estimates, a phenomenon driven by their optimization process and often characterized by highly peaked predictive distributions \cite{guo_calibration_2017}. 
This tendency is exacerbated in medical imaging due to limited and highly imbalanced datasets. 
The scarcity of annotated samples in medical cohorts limits the model's exposure to the full spectrum of pathological and anatomical variations, leading to poor generalization and reliance on spurious correlations, which the model often expresses with unjustified confidence \cite{silva_filho_classifier_2023}. 
Furthermore, unlike classification tasks, where calibration refers to aligning predicted probabilities with true outcome frequencies, segmentation problems introduce spatial and structural complexity, requiring voxel-wise confidence estimation across heterogeneous anatomical regions. 
Consequently, standard calibration techniques, effective for classification, often fail to generalize to segmentation tasks, since they neglect the spatial dependencies and coherence in a problem where contextual dependencies matter \cite{wang_calibrating_2023}. 
Adjusting a model’s confidence distribution can easily distort decision boundaries and thus degrade segmentation accuracy. 
Therefore, calibrating segmentation models without compromising their spatial precision or clinical relevance remains an open research problem \cite{huang_review_2024,lambert_trustworthy_2024}.

Prior research addressing calibration limitations in deep segmentation models can be broadly categorized into three main approaches: post-hoc recalibration, training-time methods, and techniques focusing on holistic uncertainty quantification. 
Post-hoc recalibration methods like Temperature Scaling \cite{rousseau_post_2025} are simple and do not require adjusting model parameters, but necessitate extra labeled data.
Training-time techniques like Label Smoothing \cite{bathula_deepti_r_ls_2024}, specialized loss functions \cite{galdran_multi-head_2023} or confidence-weighted training regimes \cite{dawood_uncertainty_2023} can be more effective, although they need to be integrated alongside model training, which is not always not feasible.
In addition, these often require careful tuning to balance the calibration objective against the primary segmentation accuracy. 
Third, Uncertainty Quantification (UQ) methods aim to natively model confidence by treating uncertainty as a core component of the model. 
Approaches like Bayesian Neural Networks \cite{lemay_improving_2022} or Evidential Deep Learning \cite{dawood_addressing_2023} provide estimates of epistemic uncertainty, which can be translated into improved calibration reliability. 
Despite recent advances, a universally reliable and accuracy-preserving calibration mechanism remains elusive, highlighting the need for further research.

A critical and often overlooked source of uncertainty in medical image segmentation arises not from model confidence, but from human annotation variability \cite{schmarje_is_2022}. 
Indeed, even experienced radiologists may disagree, particularly in ambiguous or low-contrast regions, making consistent and precise delineation a major challenge \cite{riera-marin_calibration_2025}. 
There is a connection underlying inter-rater variability and calibration under aleatoric (data-related) uncertainty: disagreement between expert clinicians when defining anatomical or pathological boundaries should contribute to greater predictive uncertainty, whereas annotator consensus should result in higher model confidence. 
Surprisingly, this connection has been mostly overlooked in medical image analysis research, where multi-rater annotations are often treated as label noise \cite{wang_learning_2025}, rather than as a signal for improving model calibration. 
Therefore, most previous works focus on how to optimally merge different annotations. 
This can be achieved with straightforward label fusion techniques, e.g. majority voting, weighted voting, or even random label sampling \cite{lemay_label_2023}. 
More sophisticated approaches attempt to estimate individual annotator skill \cite{zhang_learning_2023} or construct specialized label smoothing strategies \cite{zhang_soft_2023}. 
These methods do not necessarily translate to segmentation problems \cite{islam_spatially_2021}, which justifies the development of specialized techniques like STAPLE and its many variants \cite{warfield_simultaneous_2004}. 
However, to the best of our knowledge, few works have studied the link between annotation consensus and calibration, mostly limited to simple classification tasks \cite{jensen_improving_2019} or requiring complex architecture designs or training regimes \cite{ji_learning_2021}.

In this paper, we introduce a new approach to bridge the gap between model calibration and data with multiple annotators. 
Our key observation is that the underlying uncertainty arising from multiple annotations can be implicitly modeled by means of ordinal labels. 
In classification problems, ordinal labels arise when classes reflect an intrinsic ranking like severity progression (e.g., disease stages, risk levels), and misclassifications are penalized proportionally to their distance from true labels \cite{vuong_joint_2021}. 
Here we propose to interpret the spread of expert opinions as ordinal indicators of confidence, as illustrated in Fig \ref{key_figure}. 
In this way, we can naturally capture multi-expert uncertainty in a simple manner, even when dealing with spatially-varying uncertainties in segmentation tasks. 
Our approach allows the exploitation of established ordinal classification techniques that exist in the literature \cite{galdran_cost-sensitive_2020, lee_dior-vit_2025, wen_ordinal_2023} to capture annotator consensus and improve model calibration. 
To this end, we propose training segmentation models by optimizing a Ranked Probability Score (RPS) loss \cite{galdran_performance_2023}, a proper scoring rule specifically suited for ordinal alignment.
While our preliminary work introduced the MR-ECE metric as a tool for quantifying calibration in multi-rater scenarios \cite{riera-marin_multi-rater_2026}, this manuscript proposes the first methodological framework designed specifically to minimize this error. The primary contributions of this work are:
\begin{itemize}
    \item Novel Ordinal Formulation for Multi-Rater Supervision: We shift the paradigm from treating rater disagreement as independent noise to a structured ordinal hierarchy. 
    This allows the model to learn the "intensity" of consensus, which is a significant conceptual advance over existing soft-label methods.
    \item Calibration-Aware Training via RPS Loss: We introduce the use of the RPS as a regularization term for segmentation. We demonstrate that this loss is uniquely suited for multi-rater data because it penalizes probability assignments based on their distance from the empirical distribution of experts.
    \item Multi-Modal Empirical Validation: We provide the first comprehensive evaluation of this ordinal framework across four diverse biomedical datasets (Lung CT, Retinal Fundus, and Histopathology), demonstrating consistent improvements in model reliability.
    \item Robust Uncertainty Quantification: We apply and further validate a specialized calibration measure for multi-rater scenarios (MR-ECE), providing a rigorous framework for assessing model reliability in the presence of expert disagreement.
\end{itemize}

\section{Methodology}\label{sec_methodology}

\subsection{Notation and Problem Statement}
We consider binary voxel-wise segmentation. Let $V$ denote the set of voxels (pixels in 2D) in an image. For each voxel $v \in V$, the ground-truth label is $y(v) \in \{0,1\}$, where $y(v)=1$ denotes foreground and $y(v)=0$ background.

A segmentation model outputs, for each voxel $v \in V$, a foreground probability $\hat p(v) \in [0,1]$. 
A binary prediction is obtained by thresholding:
\[
\hat y(v) = \mathbb{I}\big(\hat p(v) \ge \tau\big),
\]
where $\tau \in (0,1)$ is a fixed threshold (typically $\tau = 0.5$).

A model is \emph{calibrated} if its predicted probabilities coincide with empirical frequencies. 
Specifically, for any confidence level $c \in [0,1]$,
\[
\mathbb{P}\big(y(v)=1 \mid \hat p(v)=c\big) = c.
\]
In words: among voxels assigned probability $c$ of being foreground, approximately a fraction $c$ are truly foreground.

It is important to stress that calibration and discrimination are orthogonal model attributes: discrimination is typically quantified from thresholded predictions $\hat y(v)$ using metrics such as accuracy or Dice, whereas calibration assesses the reliability of the probabilistic output $\hat p(v)$, independently of any particular decision threshold. These criteria capture different aspects of predictive quality; in particular, strong discriminative performance does not, in general, imply good calibration (and vice versa).

Miscalibration can be measured using a \emph{Calibration Error} metric.
To estimate this error, predictions must first be grouped into $M$ confidence bins $B_m$.
We define the per-bin average confidence ($\text{Conf}$) and accuracy ($\text{Acc}$) as follows:
\begin{equation*}
\begin{split}
    &\text{Conf}(B_m) = \frac{1}{|B_m|} \sum_{v\in B_m} \hat{p}(v), \\
    &\text{Acc}(B_m) = \frac{1}{|B_m|} \sum_{v\in B_m} \mathbb{I}\big(y(v)=\hat{y}(v)\big),
\end{split}
\end{equation*}
where $|B_m|$ is the number of voxels falling into bin $B_m$.
These quantities allow the definition of the Expected Calibration Error (ECE), a widely used empirical measure of miscalibration for voxel-level segmentation tasks:
\begin{equation*}
\displaystyle \text{ECE} = \sum_{m=1}^{M} \frac{|B_m|}{N} \left| \text{Acc}(B_m) - \text{Conf}(B_m) \right|,
\end{equation*}
where $N$ is the total number of voxels in the test set. The weights $\frac{|B_m|}{N}$ correspond to the empirical frequency of each bin, yielding a sample-based approximation to an expectation over the model's confidence distribution.

In practice, Calibration Error estimates can be unstable: they are sensitive to the binning scheme and to the choice of $M$, and they can exhibit substantial variance for limited test-set sizes.

The formulation above assumes a single ground-truth label per voxel and treats correctness as a binary property. 
In medical image segmentation, however, voxel labels can be inherently ambiguous due to weak boundaries, partial-volume effects, and protocol or rater-dependent delineations. 
Consequently, a single annotation may capture only one plausible ``foreground-ness explanation''.
When multiple annotations are available, inter-rater disagreement provides a direct empirical handle on this ambiguity and suggests that predicted probabilities should reflect it. 
In the following sections, we introduce a calibration-aware loss function for multi-annotator supervision and describe a corresponding calibration error estimator that extends the single-rater case.

\subsection{Ordinal Consensus Learning for Calibration}

\subsubsection{Multi-Rater Ordinal Annotation Modeling}

Voxel-level labels in medical image segmentation are often ambiguous, but when multiple expert annotations exist, inter-rater disagreement can be used as a proxy signal for aleatoric uncertainty. 
Our main hypothesis is that leveraging this signal can improve the calibration of the model's probabilistic predictions. 
In this work, we propose to reformulate binary segmentation with multi-rater supervision as an ordinal prediction task, enabling the use of ordinal classification losses for training.

Formally, each voxel $v$ is labeled by $K$ annotators, producing binary labels $\{y^{(r)}(v)\}_{r=1}^{K}$, where $y^{(r)}(v)\in\{0,1\}$ and the superscript $(r)$ denotes rater index. 
We summarize these votes with the \emph{Ordinal Rater Consensus} (ORC) label $\tilde{y}(v)$:
\begin{equation}
\tilde{y}(v)=\sum_{r=1}^{K} y^{(r)}(v), \qquad \tilde{y}(v)\in\{0,1,\ldots,K\}.
\end{equation}
Values $\tilde{y}(v)=0$ and $\tilde{y}(v)=K$ indicate full agreement on background and foreground, respectively, while intermediate values reflect partial agreement among annotators.

The ORC label $\tilde{y}(v)$ captures annotator agreement on an ordinal scale from $0$ to $K$. 
This lets us reformulate learning from multi-rater annotations as predicting the level of consensus at each voxel, treating binary segmentation as an ordinal classification problem with $K+1$ ordered categories. 
Hence, we design our segmentation model to return a probability distribution $\{\hat{p}_k(v)\}_{k=0}^{K}$ over consensus levels, where $\hat{p}_k(v)\approx \mathbb{P}(\tilde{y}(v)=k)$. 
In test time, the ordinal predictive distribution can be mapped back to a binary foreground probability by aggregating the probability mass assigned to majority foreground consensus levels:
\begin{equation}
\hat{p}(v)=\sum_{k \geq K/2}^{K}\hat{p}_k(v).\label{eq:aggregating_probs}
\end{equation}
Here, $\hat{p}_k(v)$ denotes the predicted probability of consensus level $k\in\{0,\ldots,K\}$ at voxel $v$, and $K$ is the number of annotators. 
In words, $\hat{p}(v)$ is the model's confidence that a majority of annotators would label voxel $v$ as foreground.

By modeling and predicting the distribution of annotator consensus rather than a single binary label, this formulation encourages the model’s probabilistic outputs to reflect ambiguity arising from inter-rater variability. 
At the same time, deriving a foreground confidence from majority-consensus levels preserves compatibility with standard binary segmentation objetives. Our central hypothesis is that this reformulation improves calibration with respect to annotator disagreement without sacrificing discriminative performance.

With multi-rater supervision expressed as ordinal consensus prediction, training becomes an instance of probabilistic learning with ordered labels. 
Ordinal classification commonly arises when errors have an ordered structure (e.g., disease staging), 
and a wide range of loss functions have been proposed to explicitly enforce this ordering in the predictions \cite{vuong_joint_2021, galdran_cost-sensitive_2020, lee_dior-vit_2025, wen_ordinal_2023, galdran_performance_2023}.
In our context, by treating the consensus level $\tilde{y}(v)\in\{0,\ldots,K\}$ as an ordered target, we can repurpose ordinal losses to learn distributions over annotator agreement and ultimately achieve better calibration at the binary predictive level. Among the many available choices, we use the Ranked Probability Score, described next.

\subsubsection{Ranked Probability Score Loss for Ordered Classification}
The Ranked Probability Score (RPS) is a proper scoring rule originally introduced for probabilistic forecasting of ordered outcomes, whose theoretical properties and decompositions have been extensively characterized in statistical literature \cite{arnold_decompositions_2024}.
It measures how well the predicted cumulative probabilities align with the true cumulative distribution induced by the ground-truth ordinal label.
Formally, for a voxel $v$, the RPS loss is defined as:
\begin{equation}
\pazocal{L}_{\mathrm{RPS}}=\frac{1}{K+1}\sum_{j=0}^{K}\big(F_j(v)-\hat{F}_j(v)\big)^2,
\end{equation}
where $F_j(v)$ and $\hat{F}_j(v)$ denote the cumulative distributions of the ground-truth and predicted ordinal labels, respectively:
\begin{equation}
F_j(v)=\sum_{k=0}^{j} y_k(v), \qquad \hat{F}_j(v)=\sum_{k=0}^{j}\hat{p}_k(v).
\end{equation}
Here, $\hat{p}_k(v)$ denotes the model’s predicted probability for the $k$-th ordinal category (consensus level), and $y_k(v)$ is the one-hot encoding of the ground-truth ORC label $\tilde{y}(v)$. The cumulative formulation through $F_j(v)$ and $\hat{F}_j(v)$ ensures that predictions close to the true consensus level are penalized less than predictions that are farther away, thereby explicitly respecting the ordinal structure of the target space. 
Unlike categorical cross-entropy or standard soft-labeling, which treat consensus as independent categories or a continuous scalar, the RPS loss assigns graded penalties proportional to the ordinal distance from the ground truth. 
While soft-labels ignore the intrinsic ranking of rater agreement, our approach ensures that predictions close to the true rater count are penalized less than distant ones. 
By treating consensus as a structured ordinal distribution rather than independent noise, we capture the hierarchical nature of expert disagreement that traditional methods overlook.

In practice, optimizing the RPS loss alone may lead to slow convergence or unstable gradients in the early stages of training, particularly when a large fraction of voxels belong to extreme consensus levels (i.e., full agreement or complete disagreement among annotators). 
Moreover, the ordinal objective can be demanding early on, since it requires the model to resolve fine-grained consensus levels (e.g., distinguishing whether $5$, $6$, or $7$ annotators vote for foreground) before robust spatial structure has been learned.

In order to improve training stability and preserve discriminative performance at the binary level, we complement the RPS term with a binary cross-entropy (BCE) loss. 
Specifically, we map the ordinal predictive distribution to a foreground confidence by summing the probability mass assigned to majority consensus levels, and compute BCE against the corresponding majority-vote foreground label derived from $\tilde{y}(v)$. 
In practice, this auxiliary term helps guide optimization in early training while remaining aligned with the ordinal consensus formulation.

The final training objective is thus defined as:
\begin{equation}
\pazocal{L}=\pazocal{L}_{\mathrm{BCE}}+\alpha\,\pazocal{L}_{\mathrm{RPS}},
\end{equation}
where $\alpha$ is a weighting coefficient that balances binary discrimination and ordinal-consensus learning. This hybrid objective encourages the network to retain strong binary segmentation performance while learning the graded uncertainty structure encoded by multi-rater annotations.

\subsubsection{Implementation and Technical Pipeline}
The proposed framework follows a structured three-stage pipeline to bridge multi-rater annotations with calibrated segmentation:
\begin{itemize}
    \item \textbf{Ordinal Target Generation:} For each voxel $v$, the $K$ binary expert annotations are summed to produce a discrete ordinal target $\tilde{y}(v) \in \{0, \dots, K\}$, representing the level of rater consensus.
    \item \textbf{Architecture and Output Heads:} We utilize a standard segmentation backbone (e.g., U-Net). The final layer is modified to feature $K+1$ output channels, followed by a softmax activation. This allows the network to predict a categorical probability distribution over all possible consensus levels $\{\hat{p}_k(v)\}_{k=0}^{K}$.
    \item \textbf{Hybrid Supervision and Inference:} The network is supervised by the joint $\pazocal{L}_{\mathrm{RPS}}$ and $\pazocal{L}_{\mathrm{BCE}}$ loss. During inference, the binary foreground probability $\hat{p}(v)$ is derived by summing the probability mass of majority-vote levels as defined in Eq. (\ref{eq:aggregating_probs}). This specific mapping logic ensures that the model preserves high-level spatial structure while penalizing uncertainty based on the ordinal distance from expert consensus.
\end{itemize}

\subsection{Evaluating Calibration with Multiple Annotations}
When multiple annotations are available, voxel labels may reflect both intrinsic ambiguity and rater-dependent annotation patterns, so calibration assessment against a single reference label can be misleading. 
In this setting, multiple annotations can also improve the robustness of calibration estimates by providing repeated observations per voxel. 
Motivated by this, we adopt a calibration error estimator specifically tailored to the multi-rater scenario, which we describe next.

\subsubsection{Calibration Error Analysis}

To assess model calibration under multiple annotators, we use our previously proposed multi-rater calibration framework \cite{riera-marin_multi-rater_2026}, which generalizes calibration error estimation to explicitly account for inter-rater variability. 
In contrast to conventional single-rater estimation, which compares each voxel prediction to a single reference label, our multi-rater formulation evaluates predictions against the labels provided by all annotators. 
We emphasize that MR-ECE is a mathematically grounded generalization of the widely accepted Expected Calibration Error (ECE), rather than a task-specific heuristic. 
It generalizes calibration error estimation to explicitly account for inter-rater variability in scenarios where a single, deterministic ground truth is undefined. 
By providing repeated observations per voxel, this estimator improves the robustness of calibration estimates and faithfully reflects the underlying clinical uncertainty. 
In practice, this can be seen as virtually expanding the test set via voxel-annotator pairs.
For the standard ECE, the $\text{MR-ECE}$ is formally defined as:
\begin{equation}\label{mr_ece}  
\text{MR-ECE}=\sum_{m=1}^M \frac{|B^{\text{MR}}_m|}{N_{total}} \ |\text{Conf}(B^{\text{MR}}_m) - \text{Acc}(B^{\text{MR}}_m)|,
\end{equation}
where $|B^{\text{MR}}_m|$ is the number of voxel--annotator pairs assigned to bin $B^{\text{MR}}_m$, and $N_{\text{total}}=N_{\text{voxels}}\cdot K$ is the total number of such pairs.
Here, $\text{Conf}(B^{\text{MR}}_m)$ is the mean predicted foreground probability in the bin, and $\text{Acc}(B^{\text{MR}}_m)$ is the empirical foreground frequency across raters.

The estimator in Eq. (\ref{mr_ece}) increases the effective number of samples by a factor of $K$, yielding more stable calibration estimates than in the single-rater case.
As a result, $\text{MR-ECE}$ is well-suited to medical image segmentation settings where annotator disagreement is common and test sets are limited.

\subsubsection{Discriminative Performance Analysis}
Calibration measures how well predicted probabilities reflect the likelihood of correctness, which can be essential for risk-sensitive downstream applications.
Nevertheless, assessing calibration in isolation is insufficient: a model may be well calibrated while offering limited discriminative power, and conversely, a highly discriminative model may produce systematically miscalibrated probabilities.

To complement our calibration analysis, we evaluate discriminative performance using the Area Under the Receiver Operating Characteristic Curve (AUC), which measures class separability independently of any fixed threshold.
Reporting $\text{MR-ECE}$ together with AUC provides a more complete characterization of both calibration and discriminative performance, as well as potential trade-offs between them.
While the Dice Similarity Coefficient (DSC) is a standard metric for binary segmentation, its application in multi-rater settings is often problematic as it requires collapsing the expert distribution into a single "gold standard" via consensus (e.g., majority vote or STAPLE). 
Such an approach is fundamentally at odds with our objective of preserving and calibrating the full spectrum of rater disagreement. 
Because DSC is a threshold-dependent overlap metric, it fails to capture the reliability of probabilistic predictions, which is the central focus of this study. 
Consequently, we prioritize threshold-independent measures like AUC and MR-ECE, which evaluate the model’s ability to represent inherent ambiguity without the information loss associated with binary consensus-averaging.

\begin{figure*}[!t]
\centering
\includegraphics[width=0.94\textwidth]{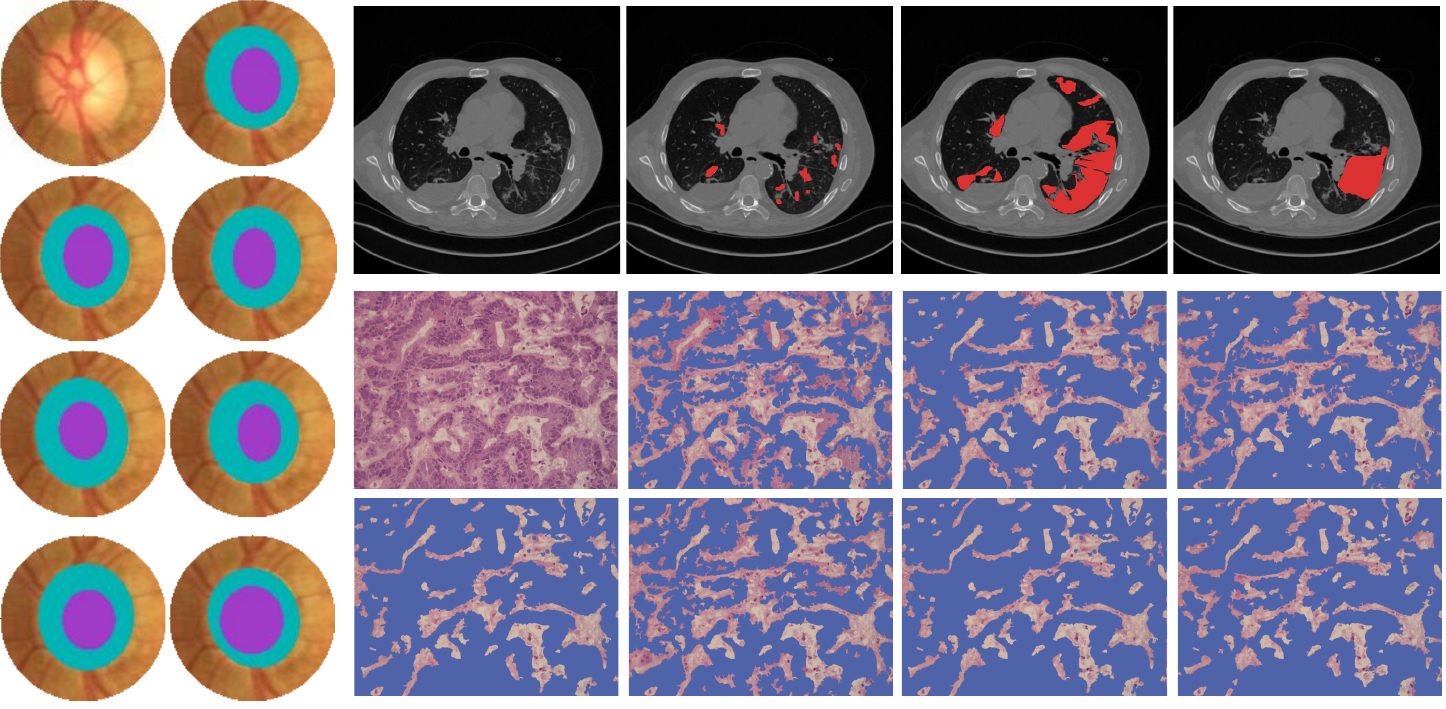} 
\caption{Representative samples and multi-rater annotations from the segmentation datasets used to evaluate the proposed ordinal agreement strategy. Left: REFUGE, retinal fundus image with optic disc (OD, outer boundary) and optic cup (OC, inner boundary) segmentations provided by multiple expert annotators. Top: LongCIU, axial chest CT slice with expert annotations of ground-glass opacities (GGO) associated with post-acute Long COVID. Bottom: CoCaHis, histopathological colon tissue image with pixel-wise cancer region annotations from multiple pathologists. The figure highlights the modality and task-dependent nature of annotation variability across datasets.} 
\label{example_fusion}
\end{figure*}
\section{Experimental Analysis}\label{sec_experimental_results}


\subsection{Datasets}
We conduct our validation on several binary segmentation problems with multiple independent ground-truth annotations per sample, as illustrated in Fig.~\ref{example_fusion}.
We give a detailed description of each of the considered datasets below.

\begin{itemize}[leftmargin=*, topsep=2pt, itemsep=2pt, parsep=2pt]
\item \textbf{CoCaHis (Colon Cancer Histopathological Dataset)} \cite{sitnik_dataset_2021} is a publicly available dataset created to support the development and evaluation of AI models in histopathological image analysis for colorectal cancer. 
It consists of 82 high-resolution whole slide images of colon tissue samples, annotated by seven pathologists to indicate cancerous and non-cancerous regions. 
The dataset serves as a benchmark for colorectal cancer segmentation with substantial inter-rater variability.


\item \textbf{REFUGE (Retinal Fundus Glaucoma Challenge)} \cite{orlando_refuge_2020} is a benchmark dataset designed to facilitate the development of automated methods for glaucoma detection and optic nerve head analysis from color fundus images.
Introduced through a MICCAI challenge, REFUGE contains 1,200 high-resolution retinal fundus images annotated by trained ophthalmologists.
Each image is accompanied by seven independent segmentations of both the optic disc and the optic cup, which we treat as two distinct binary segmentation tasks.
While optic disc delineation is relatively well-defined and exhibits moderate inter-rater variability, optic cup segmentation is considerably more challenging, with substantially higher annotation ambiguity due to weaker boundaries and lower contrast.
The dataset spans images acquired from two different devices (Zeiss and Canon), introducing additional appearance variability across both tasks.

\item The \textbf{Long COVID Iowa‑UNICAMP dataset (LongCIU)} \cite{carmo_long_2024} is the first publicly available CT image collection focusing on post-acute lung pathology in Long COVID patients. 
It comprises 90 anonymized axial CT slices, each manually segmented by three independent expert annotators for ground-glass opacities (GGO) and consolidations, resulting in 360 annotated masks. 
The CT images were acquired at the University of Iowa Hospitals with a standardized non-contrast protocol (1mm slice thickness) from adults still symptomatic 30+ days after COVID‑19 infection. 
The dataset provides per-annotator masks together with a STAPLE-based consensus, enabling explicit analysis of inter-rater variability. 
\end{itemize}
Across these datasets, the availability of multiple independent annotations per sample allows us to study calibration under varying degrees of inter-rater disagreement across modalities.

\begin{table*}[t]
\centering
\renewcommand{\arraystretch}{1.25}
\setlength{\tabcolsep}{12pt}
\small
\caption{MR-ECE across the four benchmark datasets for different annotation handling strategies.
Lower values indicate better calibration with respect to multi-rater agreement.
MR-ECE values are scaled by $10^{2}$ for readability.}
\label{tab:mr_ece}

\begin{tabular}{lcccc}
\toprule
\textbf{Method} & \textbf{CoCaHis} & \textbf{REFUGE-DISC} & \textbf{REFUGE-CUP} & \textbf{LongCIU} \\
\midrule
Random Sampling      & $3.39 \,\pm\, 0.98$ & $3.41 \,\pm\, 0.24$ & $12.01 \,\pm\, 0.99$ & $5.17 \,\pm\, 0.38$ \\
Median Consensus     & $5.68 \,\pm\, 0.47$ & $3.53 \,\pm\, 0.12$ & $13.29 \,\pm\, 0.82$ & $6.19 \,\pm\, 0.24$ \\
\addlinespace
Soft Consensus      & $9.23 \,\pm\, 1.82$ & $6.00 \,\pm\, 0.24$ & $20.39 \,\pm\, 0.89$ & $5.69 \,\pm\, 0.27$ \\
Soft Consensus-G    & $11.34 \,\pm\, 2.10$ & $7.95 \,\pm\, 0.26$ & $24.31 \,\pm\, 0.80$ & $6.97 \,\pm\, 0.35$ \\
\addlinespace
SIMPLE               & $14.88 \,\pm\, 1.32$ & $3.54 \,\pm\, 0.12$ & $13.39 \,\pm\, 0.78$ & $6.56 \,\pm\, 0.24$ \\
STAPLE               & $5.50 \,\pm\, 0.71$ & $4.06 \,\pm\, 0.14$ & $16.39 \,\pm\, 1.01$ & $7.93 \,\pm\, 0.26$ \\
SVLS                 & $5.28 \,\pm\, 0.63$ & $3.48 \,\pm\, 0.14$ & $13.21 \,\pm\, 0.89$ & $6.15 \,\pm\, 0.22$ \\
\addlinespace
\textbf{RPS (ours)}  & $\mathbf{2.41 \,\pm\, 0.40}$ & $\mathbf{1.09 \,\pm\, 0.15}$ & $\mathbf{6.67 \,\pm\, 0.35}$ & $\mathbf{1.48 \,\pm\, 0.26}$ \\
\bottomrule
\end{tabular}
\end{table*}

\subsection{Experimental Protocol}
In this section, we evaluate different multi-annotation handling strategies across all datasets under a common experimental setup.
All models are implemented using a standard encoder--decoder architecture with a standard pretrained ResNet backbone and a Feature Pyramid Network decoder, based on the \texttt{segmentation\_models\_pytorch} library.

\subsubsection{Multi-Annotation Handling Strategies}
Handling multiple annotations in medical image segmentation requires strategies that account for inter-rater variability while providing a usable supervision signal.
In our experiments, we evaluate several established multi-annotation handling strategies for comparison with our proposed ordinal learning approach.

\textbf{Classical Strategies:} 
We consider commonly used hard fusion baselines that reduce multiple annotations to a single training label.
Simple extremes such as voxel-wise intersection or union were excluded from this study, as these methods have been extensively investigated and shown to be suboptimal for generating representative training labels \cite{warfield_simultaneous_2004}.
\begin{itemize}[label=--, leftmargin=*, topsep=2pt, itemsep=2pt, parsep=2pt]
    \item Random Sampling (RS): For each training iteration, the label map from a single annotator is selected uniformly at random. RS preserves annotator-specific variability but provides a stochastically regularized supervision signal \cite{jensen_improving_2019}.
    \item Median Consensus (MC): The per-voxel median across annotations is computed to produce a hard segmentation mask. MC suppresses outlier annotations but partially discards uncertainty information \cite{ji_learning_2021}.
\end{itemize}

\textbf{Soft Consensus Strategies:}
These approaches retain annotator disagreement via soft per-voxel supervision signals.
\begin{itemize}[label=--, leftmargin=*, topsep=2pt, itemsep=2pt, parsep=2pt]
    \item Soft Consensus (SC):
    Per-voxel probabilities are computed by averaging binary annotations across raters, yielding continuous labels in $[0,1]$ that explicitly encode annotator disagreement \cite{lourenco-silva_using_2022}.
    \item Soft Consensus with Gaussian Smoothing (SC-G):
    The SC label map is regularized using a Gaussian convolution to impose local spatial consistency and mitigate annotation noise \cite{wei_smooth_2021}.
    The smoothing kernel width $\sigma$ controls the degree of spatial uncertainty modeling.
\end{itemize}

\textbf{Probabilistic Modeling Strategies:}
\begin{itemize}[label=--, leftmargin=*, topsep=2pt, itemsep=2pt, parsep=2pt]
    \item STAPLE:
    Simultaneous Truth And Performance Level Estimation iteratively estimates a latent ground-truth segmentation while modeling annotator-specific sensitivity and specificity \cite{warfield_simultaneous_2004}.
    \item SIMPLE:
    The Selective and Iterative Method for Performance Level Estimation extends STAPLE by incorporating additional structural regularization during label fusion \cite{langerak_label_2010}.
\end{itemize}

\textbf{Spatially Adaptive Strategies:}
This category captures methods that adapt spatial smoothing based on local annotator agreement.
\begin{itemize}[label=--, leftmargin=*, topsep=2pt, itemsep=2pt, parsep=2pt]
    \item Spatially Varying Label Smoothing (SVLS):
    Annotator agreement is used to modulate the extent of spatial smoothing applied to soft consensus labels, with higher disagreement leading to stronger smoothing \cite{islam_spatially_2021}.
\end{itemize}

\subsubsection{Hyperparameter Selection}
Hyperparameters for baseline multi-annotation handling strategies were selected based on validation set performance.
For methods involving spatial smoothing, including SVLS, we evaluated a small set of Gaussian kernel widths $\sigma \in \{0.5, 1.0, 2.0, 3.0\}$ and fixed $\sigma=1$ for all subsequent experiments.
For our proposed ordinal learning approach, we additionally examined the sensitivity of the combined training objective to the weighting coefficient $\alpha$ controlling the contribution of the Ranked Probability Score (RPS) loss.
The parameter $\alpha$ was varied in the range $[0.5, 1.0]$ in steps of $0.1$, and $\alpha=0.8$ was selected based on validation performance as a suitable trade-off between calibration and discriminative accuracy.
This value was used in all comparative experiments.

Final performance metrics were then computed on an independent test set that remained untouched during model development, ensuring an unbiased evaluation of generalization performance.

\begin{table*}[t]
\centering
\renewcommand{\arraystretch}{1.25}
\setlength{\tabcolsep}{12pt}
\small
\caption{AUC across the four benchmark datasets for different annotation handling strategies.
Higher values indicate better discriminative performance. Reported values are mean $\pm$ standard deviation.}
\label{tab:auc}

\begin{tabular}{lcccc}
\toprule
\textbf{Method} & \textbf{CoCaHis} & \textbf{REFUGE-DISC} & \textbf{REFUGE-CUP} & \textbf{LongCIU} \\
\midrule
Random Sampling      & $95.47 \,\pm\, 4.70$ & $99.13 \,\pm\, 0.83$ & $92.06 \,\pm\, 7.03$ & $92.53 \,\pm\, 4.67$ \\
Median Consensus     & $95.10 \,\pm\, 3.76$ & $99.06 \,\pm\, 0.54$ & $90.84 \,\pm\, 7.38$ & $\mathbf{94.32 \,\pm\, 2.73}$ \\
\addlinespace
Soft Consensus       & $94.51 \,\pm\, 5.35$ & $99.08 \,\pm\, 0.80$ & $90.15 \,\pm\, 5.82$ & $91.13 \,\pm\, 3.07$ \\
Soft Consensus-G    & $94.16 \,\pm\, 5.87$ & $98.89 \,\pm\, 1.04$ & $89.87 \,\pm\, 6.44$ & $90.06 \,\pm\, 2.77$ \\
\addlinespace
SIMPLE               & $95.11 \,\pm\, 5.51$ & $99.02 \,\pm\, 0.56$ & $90.59 \,\pm\, 6.83$ & $92.98 \,\pm\, 3.06$ \\
STAPLE               & $94.82 \,\pm\, 4.43$ & $98.81 \,\pm\, 0.61$ & $89.72 \,\pm\, 8.54$ & $91.03 \,\pm\, 3.86$ \\
SVLS                 & $\mathbf{95.50 \,\pm\, 4.34}$ & $99.07 \,\pm\, 0.55$ & $90.91 \,\pm\, 7.81$ & $93.74 \,\pm\, 3.00$ \\
\addlinespace
\textbf{RPS (ours)}  & $95.05 \,\pm\, 3.75$ & $\mathbf{99.20 \,\pm\, 0.41}$ & $\mathbf{92.57 \,\pm\, 3.35}$ & $93.44 \,\pm\, 2.64$ \\
\bottomrule
\end{tabular}
\end{table*}

\subsection{Quantitative Results}
Quantitative results across all datasets are summarized in Table~\ref{tab:mr_ece} and Table~\ref{tab:auc}, reporting calibration performance using MR-ECE and discriminative performance using AUC, respectively.
To ensure the reliability of our comparative analysis and to account for potential variability in the test data, we evaluate model performance through Bootstrap Resampling ($N=10$) on the hold-out test set. 
In each iteration, we randomly sample 60\% of the test set with replacement to estimate the empirical distribution of our primary metrics (AUC and MR-ECE). 
This approach allows us to report not only the mean performance but also robust confidence intervals, confirming that the observed improvements in calibration are statistically consistent across diverse data subsets rather than artifacts of a specific test-set configuration. By characterizing the evaluation uncertainty in this manner, we provide a more rigorous assessment of model behavior in the presence of multi-rater labels.

Table~\ref{tab:mr_ece} shows that the proposed RPS-based ordinal framework consistently achieves the lowest MR-ECE across all datasets, indicating superior calibration with respect to multi-rater agreement.
In contrast, classical hard fusion strategies such as Median Consensus and probabilistic averaging methods such as Soft Consensus exhibit substantially higher calibration error and variability.
While methods such as STAPLE, SIMPLE, and SVLS improve calibration in some settings, their performance remains less consistent across datasets, particularly in the presence of pronounced inter-rater disagreement.

Discriminative performance, reported in Table~\ref{tab:auc}, reveals a different pattern.
Several baseline methods, including Random Sampling and Median Consensus, achieve competitive AUC values across datasets, suggesting that strong discrimination can be obtained even with simple annotation handling strategies.
Importantly, the proposed RPS framework matches or exceeds the best-performing baselines in terms of AUC on REFUGE-OC and REFUGE-OD, while maintaining competitive performance on CoCaHis and LongCIU.
These results indicate that the calibration gains achieved by ordinal-aware training are not obtained at the expense of discriminative power.

Taken together, the quantitative results demonstrate that explicitly modeling ordinal consensus via the RPS loss provides a favorable balance between calibration and discrimination.
Compared to existing annotation handling strategies, the proposed approach yields more reliable probabilistic predictions while preserving strong classification performance across diverse imaging modalities.

\begin{figure*}[!t]
\centering
\includegraphics[width=0.95\textwidth]{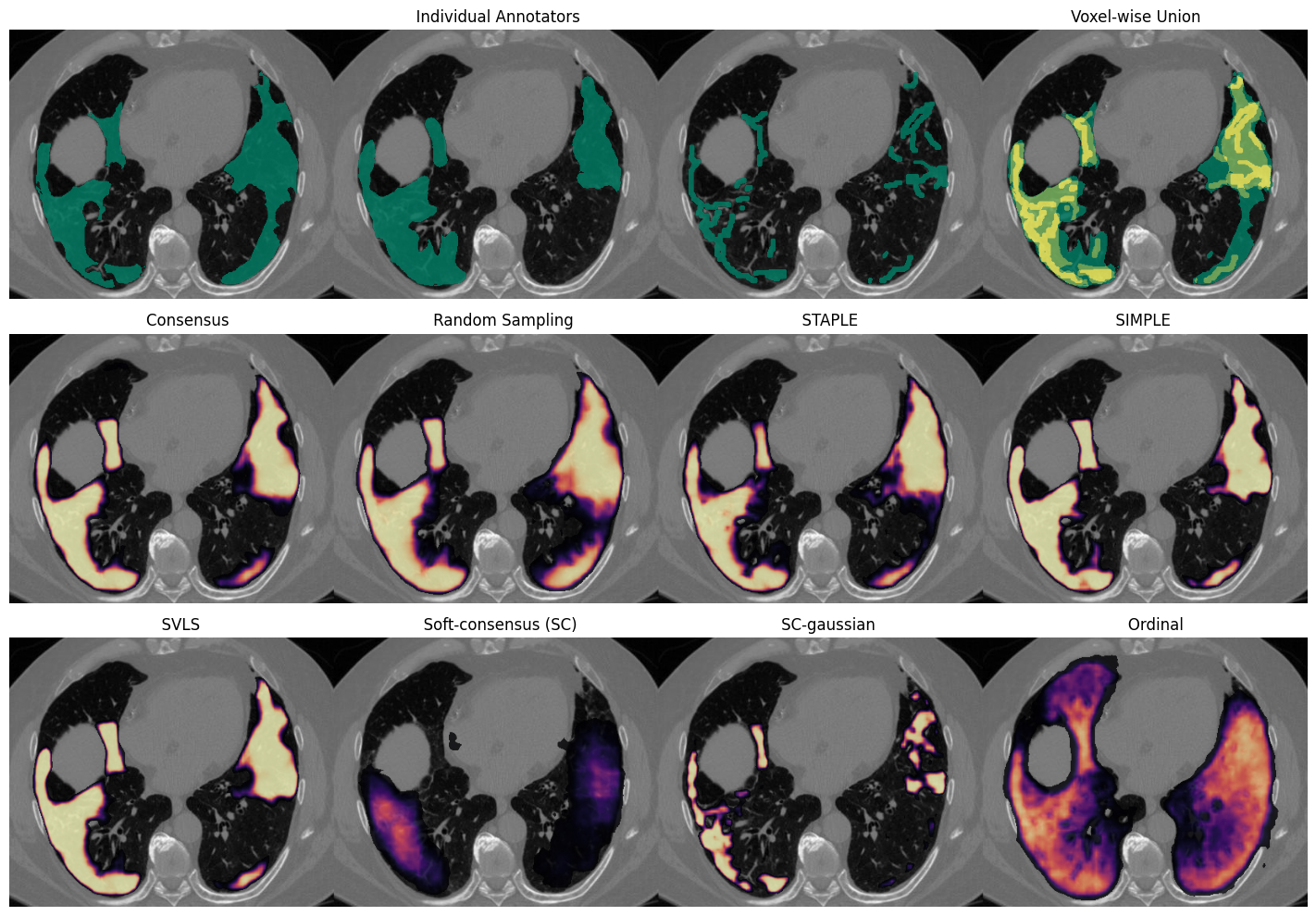} 
\caption{Qualitative comparison of segmentation calibration across different methodologies on the LongCIU dataset. Top row: Per-expert annotation and voxel-wise union, where yellow indicates full consensus and dark green indicates a single-rater annotation. Center bottom rows: Predictions from models trained using standard label fusion: Consensus, Random Sampling, STAPLE, and SIMPLE. Third row: Predictions from models utilizing uncertainty-aware strategies: SVLS, Soft-Consensus, SC-Gaussian, and our Ordinal-RPS model. Our approach is the only one capturing strong disagreement beyond annotation borders while maintaining high structural fidelity.}
\label{visual_example}
\end{figure*}

\subsection{Qualitative Analysis}
Fig.~\ref{visual_example} presents a qualitative comparison between individual expert annotations (top row) and model predictions obtained using different annotation handling strategies (bottom rows) on a representative LongCIU slice.
The top row additionally shows the voxel-wise union of annotations, encoding the degree of inter-rater agreement: voxels annotated by all experts appear in yellow, while voxels marked by a single annotator are shown in dark green.
This visualization highlights the substantial spatial variability present even among expert delineations and motivates the need for models that explicitly represent annotation uncertainty.

Conventional consensus-based and probabilistic fusion strategies, including hard consensus, Random Sampling, STAPLE, and SIMPLE, tend to produce smooth and homogeneous predictions that approximate an average annotation.
While these methods yield visually clean segmentations, they largely suppress regions of disagreement, resulting in probability maps that are effectively near-binary.
As a consequence, ambiguous regions—clearly visible in the annotation union—are assigned overly confident predictions, masking the true extent of inter-rater variability.

Methods designed to preserve uncertainty exhibit a different failure mode.
Soft Consensus (SC) produces diffuse probability maps due to excessive averaging, leading to spatially blurred predictions that dilute localized agreement patterns and appear under-confident.
The addition of Gaussian smoothing (SC-G) further amplifies this effect by prioritizing spatial regularity, often restricting confident predictions to regions of strong agreement while effectively discarding areas of partial consensus.
Similarly, SVLS tends to generate sharp segmentations that resemble hard-consensus outputs, failing to faithfully reflect the full distribution of expert disagreement despite its variance-aware design.

In contrast, the proposed ordinal framework produces reliability maps that closely mirror the spatial distribution of annotator agreement.
Ambiguous regions are explicitly highlighted through graded confidence levels rather than being collapsed into hard foreground or eliminated through over-smoothing.
By modeling consensus as an ordered structure, the ordinal approach captures disagreement beyond annotation boundaries while preserving anatomical coherence.
This enables a balanced representation of uncertainty that avoids both overconfident binarization and excessive diffusion, resulting in predictions that more accurately reflect expert variability.
Visual differences on Fig.~\ref{visual_example} are consistent with the quantitative MR-ECE improvements reported in Table~\ref{tab:mr_ece}, particularly on LongCIU, where inter-rater variability is highest.

\section{Discussion and Conclusions}\label{sec_experimental_results}

This work addresses the problem of learning calibrated probabilistic segmentation models from multiple expert annotations.
Rather than collapsing rater disagreement into a single hard target, we reformulate multi-rater supervision as an ordinal prediction task and optimize a Ranked Probability Score (RPS) loss to explicitly model graded levels of agreement.
Across four benchmark datasets with varying degrees of annotation ambiguity, this formulation consistently produced better-calibrated predictions, as reflected by substantially lower MR-ECE values, while maintaining competitive or superior discriminative performance. 

Regarding model reliability, MR-ECE represents a mathematically necessary extension of ECE that accounts for the spatial variability inherent in multi-rater settings. 
By evaluating predictions against the empirical rater distribution, it provides a more rigorous assessment of clinical uncertainty. 
The substantial reduction in MR-ECE, achieved without sacrificing discriminative power, validates our ordinal approach as an effective solution for calibration in ambiguous scenarios, even when gains in AUC are marginal.

A key outcome of this study is that improved calibration does not require sacrificing accuracy.
Quantitative results demonstrate that the ordinal–RPS framework matches or exceeds the AUC of commonly used consensus-based and probabilistic fusion strategies, while qualitative analysis reveals a more effective separation between confident and ambiguous regions.
In contrast to hard consensus methods, which tend to produce overconfident near-binary outputs, and soft averaging approaches, which often dilute localized agreement through excessive smoothing, the ordinal formulation preserves spatial structure while faithfully reflecting inter-rater variability.
This balance is particularly evident in challenging settings such as the REFUGE optic cup and LongCIU datasets, where annotation disagreement is pronounced.

Importantly, the proposed approach should be viewed as a general strategy for learning calibrated probabilistic predictions from multiple annotators by exploiting the ordinal structure implicit in multi-rater agreement.
As a consequence, our framework is not tied to a specific ordinal loss function; while we instantiated it using the Ranked Probability Score due to its properness and suitability for ordered outcomes, many alternative ordinal-aware losses could be employed, including cumulative link models, CORAL/CORN-style objectives, Wasserstein-based losses, or unimodality-enforcing regularizers.
Exploring these alternatives may further improve optimization behaviour or calibration properties, and thus represents a natural extension of this work.

Extending the proposed framework beyond binary segmentation remains an open challenge.
In multi-class settings, annotator disagreement cannot be summarized by a single ordered consensus variable, as uncertainty may arise both within and across semantic classes.
Defining meaningful ordinal structures in this context, or combining multiple binary or one-vs-rest ordinal formulations, is not straightforward and opens the door to further investigation.
Addressing these challenges will be necessary to apply ordinal calibration principles to more complex segmentation tasks.

From a practical standpoint, our findings suggest that explicitly modeling annotator disagreement during training can lead to more trustworthy probabilistic outputs, without requiring modifications to the underlying model architecture, \textit{e.g.} specialized multi-head designs. By aligning predicted probabilities with empirical agreement patterns, ordinal-aware calibration yields probability estimates that can be more reliably used in downstream risk-sensitive decision-making.

In summary, this work demonstrates that treating multi-rater annotations as structured ordinal information, rather than noise to be averaged away, provides a principled route to improved calibration. 
By reframing rater disagreement as an ordinal problem, we offer a plug-and-play methodology that integrates seamlessly into existing architectures to enhance the reliability of medical AI. 
Ultimately, we hope this perspective encourages the broader adoption of ordinal modeling and calibration-aware practices within the medical image analysis community.

\section*{Acknowledgment}
This work was supported by the Catalan Government through the Industrial Doctorates program (AGAUR 2021-063), in collaboration with Sycai Technologies SL. It is part of the CPP project (CPP2021-008364), funded by \texttt{MCIN/AEI/10.13039/501100011033} and co-financed by the European Union (NextGenerationEU/PRTR). A.G. is supported by the RYC2022-037144-I grant, funded by \texttt{MCIN/AEI/10.13039/501100011033} and co-financed by FSE+.


\end{document}